\documentclass{article}
\usepackage{iclr2027_conference,times}
\usepackage{hyperref}
\usepackage{url}
\usepackage{booktabs}
\usepackage{graphicx}
\usepackage{amsmath}
\usepackage{amssymb}
\usepackage{amsthm}

\title{Perturbing Inputs Improves\\Pretraining Distillation}
\author{Ziyang Cai$^{1}$\quad Yihe Dong$^{1}$\quad Dimitris Papailiopoulos$^{2}$\quad Shivam Garg$^{2}$\\
$^{1}$Princeton University\qquad $^{2}$Microsoft Research\\
\small\texttt{\{zc5794,ydong\}@princeton.edu}, \texttt{\{dimitriosp,shigarg\}@microsoft.com}}

\iclrfinalcopy
\begin{document}
\maketitle
\lhead{}
\renewcommand{\headrulewidth}{0pt}
\thispagestyle{fancy}

\begin{abstract}
Knowledge distillation trains a student language model to match a teacher's predictions on the training
text. We ask whether learning the teacher's behavior beyond the training text can further improve the
student. We investigate this using three input perturbations: additive token noise, blending two
documents, and blending neighboring tokens. Specifically, for an early part of pretraining we perturb the
inputs and train the student to match the teacher on them, and we spend the rest of the same token budget
on ordinary distillation. When pretraining from scratch, the best method improves on standard distillation
at both scales we tried, by 0.025 nats at 135M parameters and by 0.020 nats at 0.5B (mean of three seeds). At 135M
this is 1.45 times the gain that distillation itself provides over next-token training. We also observe
promising downstream transfer: at 0.5B, the best variant improves average zero-shot accuracy by about one
percentage point across seven tasks. It also retains its validation-loss advantage after low-bit weight
quantization. We also test a data-limited setting in which training data must be reused, and find that
the gains persist. Finally, ablations suggest that, for additive noise, perturbing the teacher's inputs
contributes most of the gain, while perturbing only the student's inputs yields much less.
\end{abstract}

\section{Introduction}

When pretraining small language models, it is common to use distillation from a larger model
\citep{hinton2015distilling,gemmateam2024gemma2,peng2025pretraining,museglimmer2026}. The student learns to
match the teacher's next-token distribution, but only on the training text itself. The teacher defines a
prediction function beyond those inputs, and its behavior elsewhere is information the student never
receives. In this work, we ask whether learning some of that behavior can improve pretraining distillation.

This question connects naturally to data augmentation. In language-model pretraining, augmentation
typically corrupts the student's input, for example by masking or replacing tokens, while keeping the
original next token as the target \citep{xu2026dataconstrained,chen2026augmentation,ni2025superdata}. A
teacher makes a different kind of augmentation possible: it can predict on the corrupted input itself,
providing a target that reflects what the student actually sees. Training on teacher-labeled augmented
inputs is established in task-specific distillation of language models
\citep{jiao2020tinybert,rashid2021matekd} and in vision \citep{beyer2022patient,wang2022goodaug}. Whether
it helps during pretraining from scratch is less clear, because training often proceeds without repeating
the data, and the teacher's predictions on clean text already provide rich supervision.

\begin{figure}[t]
\centering
\begin{minipage}[b]{0.5\linewidth}\centering
\includegraphics[width=\linewidth]{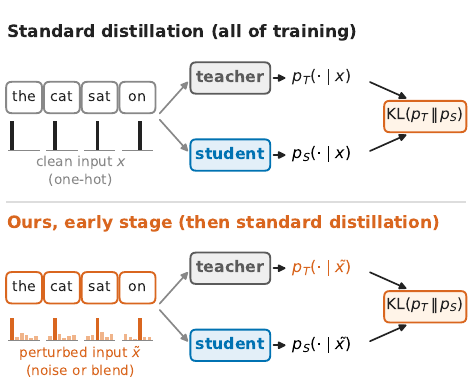}\\[-2pt]
{\small (a) Standard distillation and our perturbed stage}
\end{minipage}\hfill
\begin{minipage}[b]{0.48\linewidth}\centering
\includegraphics[width=\linewidth]{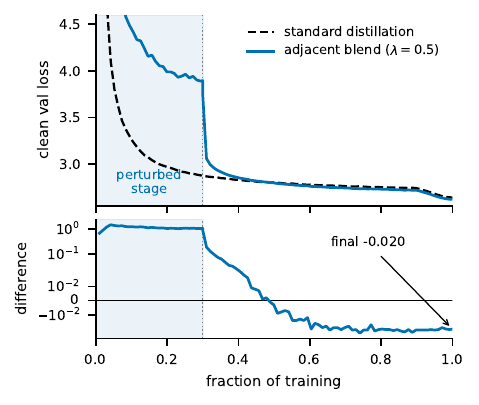}\\[-2pt]
{\small (b) Two-stage schedule (0.5B)}
\end{minipage}
\caption{\textbf{Overview.} (a) In standard distillation, teacher and student read the same clean text and
the student is trained to match the teacher's prediction on it. In the early stage of our method, both read
a perturbed version of the text, and the student matches the teacher's prediction on that perturbed input:
the inputs on which the teacher supervises the student change, not only the student's input. The bars
under each token sketch its input vector over a toy vocabulary, one-hot for clean text and spread over
several vocabulary items when perturbed (Fig.~\ref{fig:families}). The rest of
training is standard distillation on clean text. (b) Clean validation loss of two 0.5B students trained from the
same initialization on the same data order, one with standard distillation and one with adjacent blending
($\lambda{=}0.5$) for the first 30\% of training, and their difference (symmetric log scale). While its
inputs are perturbed, the blended student lags by about one nat. After the switch it recovers quickly,
overtakes the baseline at about 47\% of training and finishes 0.020 nats below it.}
\label{fig:overview}
\end{figure}

We investigate this question in pretraining from scratch through a two-stage distillation schedule
(Section~\ref{sec:method}, Fig.~\ref{fig:overview}). Specifically, we study three perturbations: additive
token noise, blending two documents position by position, and blending neighboring tokens into a single
input position. Teacher and student receive perturbed inputs during an early stage of training, followed by
ordinary distillation on clean inputs. All three perturbations improve on standard distillation at both
student scales we study (Sections~\ref{sec:setup}--\ref{sec:results}). Adjacent-token blending is the
strongest variant, reducing validation loss by 0.025 nats at 135M parameters and 0.020 nats at 0.5B, with
the same number of training steps as standard distillation. To put this improvement in context, at 135M
the additional gain is 1.45 times the gain that distillation itself provides over next-token training.

To understand the source of these gains, we run several ablations (Section~\ref{sec:ablation}). Without a
teacher, perturbing the inputs while training the student to predict the original next token provides
little benefit or is sometimes harmful. With additive noise, we also perturb only the teacher: the student
sees clean text but learns to match the teacher's prediction on a perturbed version of it. This achieves
about 90\% of the gain obtained when both models are perturbed. In contrast, perturbing only the student
while keeping the teacher's input clean achieves about a third. Thus, teacher-side perturbation
contributes most of the improvement. Further ablations find that averaging four noisy teacher predictions
gives a smaller improvement than using a single draw, despite their identical expected targets
(Appendix~\ref{app:mech}). These ablations suggest a plausible explanation: noisy teacher targets make some
fine-grained, unstable details of the teacher's predictions harder to fit precisely during training. This
may encourage the student to learn the teacher's more stable patterns without having to reproduce every
detail of its predictions on clean text.

We also study data-limited pretraining, where training repeatedly uses the same text and regularization
becomes especially important (Section~\ref{sec:dc}). With 16 epochs over 262M unique tokens, additive noise
still improves distillation by around 0.020--0.023 nats across two seeds, showing that the benefits
persist beyond the single-pass setting.

Finally, we evaluate downstream transfer and post-training weight quantization
(Section~\ref{sec:analysis}). At 135M, changes in average zero-shot accuracy are small, at most 0.6
points. At the larger 0.5B scale, the best method increases average zero-shot accuracy by about one
percentage point across seven tasks. It also retains its validation-loss advantage after low-bit weight
quantization. Together, these results suggest that perturbed-input distillation can improve the downstream
utility of compact students, rather than only their full-precision validation loss.

Our main contributions are:
\begin{itemize}
\item A proof of concept: learning a teacher's behavior beyond ordinary token inputs can improve
  language-model pretraining distillation.
\item An investigation of the source of the gains: ablations separate teacher-side and student-side effects
  and probe the role of variability in the teacher's targets.
\item Benefits beyond the main training setting: improvements persist under data repetition with tuned
  regularization, transfer to zero-shot tasks, and survive low-bit weight quantization.
\end{itemize}

\section{Method}
\label{sec:method}

\begin{figure}[t]
\centering
\includegraphics[width=\linewidth]{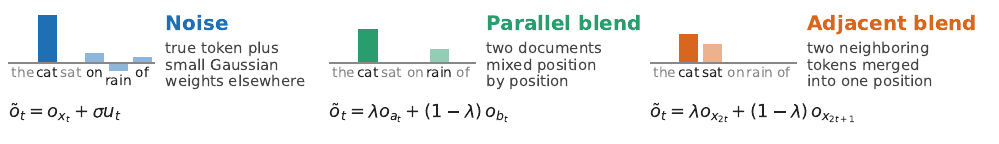}
\caption{\textbf{Perturbation families} on a toy vocabulary. Each replaces the one-hot vector of a token
with a sparse weighted combination of vocabulary items, with the position's own token shown dark. Teacher
and student both map the perturbed vector through their own embedding tables, and the student is trained
to match the frozen teacher's next-token distribution on it.}
\label{fig:families}
\end{figure}

\paragraph{One-hot perturbations (Fig.~\ref{fig:families}).} Let $o_v\in\mathbb{R}^{|V|}$ be the one-hot vector of
vocabulary item $v$, so position $t$ of a clean sequence $x$ holds $o_{x_t}$. A perturbation replaces $o_{x_t}$ by a sparse weighted combination
$\tilde o_t=\sum_j w_j\,o_{v_j}$ of a few vocabulary items. Each model computes its input embedding
as $E^\top\tilde o_t$ with its own embedding matrix $E$. Teacher and student receive the same
$(v_j,w_j)$, so they see the same perturbation whatever their embedding width. The loss is the
forward KL divergence from the teacher's next-token distribution to the student's at every
position (temperature 1).

\paragraph{Perturbation families.}
\begin{itemize}
\item \textbf{Noise.} $\tilde o_t=o_{x_t}+\sigma u_t$, where $u_t$ places independent Gaussian weights
  $\mathcal{N}(0,1/k)$ on $k{=}8$ tokens drawn uniformly from the vocabulary (with replacement, so the
  true token can occasionally be among them), resampled at every position; the weights are not
  normalized. $\sigma$ sets the size of the perturbation relative to the true token.
\item \textbf{Parallel blend.} Two independent documents $a,b$ are mixed position by position,
  $\tilde o_t=\lambda o_{a_t}+(1-\lambda)o_{b_t}$.
\item \textbf{Adjacent blend.} Consecutive token pairs of one document are merged into one position,
  $\tilde o_t=\lambda o_{x_{2t}}+(1-\lambda)o_{x_{2t+1}}$. Sequences are drawn twice as long, so
  inputs keep their full length.
\end{itemize}
The blends also pack two corpus tokens into each model-input position during the perturbed stage,
increasing text exposure without increasing the model sequence length or the number of training steps.
Outside Section~\ref{sec:dc}, all token budgets are sized so that no token is seen twice. We test whether the extra text matters in
Section~\ref{sec:ablation}.

\paragraph{Two-stage schedule.} Training consists of a perturbed stage covering a fixed fraction of
the run, followed by ordinary distillation on clean inputs for the remainder
(Fig.~\ref{fig:overview}b). The learning rate follows a warmup--stable--decay schedule
\citep{hu2024minicpm}: 10\% warmup, a constant plateau spanning the stage boundary, and a 10\%
final decay, which therefore always takes place on clean inputs. The baseline is the identical run
without the perturbed stage.

\section{Experimental setup}
\label{sec:setup}

\paragraph{Models and data.} \emph{135M scale:} a SmolLM2-135M architecture trained from random
initialization and distilled from SmolLM2-1.7B \citep{allal2025smollm2}, for 8000 steps of
$512\times1024$ tokens (4.2B tokens) on a 58.6B-token FineWeb-Edu corpus \citep{penedo2024fineweb}.
\emph{0.5B scale:} a Qwen2.5-0.5B architecture trained from random initialization and distilled
from Qwen2.5-1.5B \citep{qwen2025qwen25}, for 19073 steps of $256\times2048$ tokens (10B tokens) on
a 65B-token FineWeb-Edu corpus. No data is repeated at either scale, except in the data-constrained
experiments of Section~\ref{sec:dc}.

\paragraph{Paired comparisons.} All runs at a scale read the same data in the same order and start
from the same initialization, and differ only in the training objective. As a check, runs from the
same seed agree on validation loss at the first evaluation to within 0.004 nats. The next-token-prediction
reference uses the same data stream.

\paragraph{Evaluation.} At 0.5B every final model is scored on the full held-out validation split
(2441 non-overlapping windows of 2048 tokens, identical for every run). We report the difference to
standard distillation paired per window, with its standard error. At 135M we report the
validation loss from the training run's evaluation pass; differences below about 0.003 nats are
within its resolution. Most configurations are trained once. The main 135M ablations, the
data-constrained comparison and the best 0.5B configuration are repeated with two or three initialization
seeds, always on the same data order. For configurations trained once, error bars reflect evaluation
sampling rather than variation between training runs.

\section{Results}
\label{sec:results}

\begin{figure}[t]
\centering
\includegraphics[width=\linewidth]{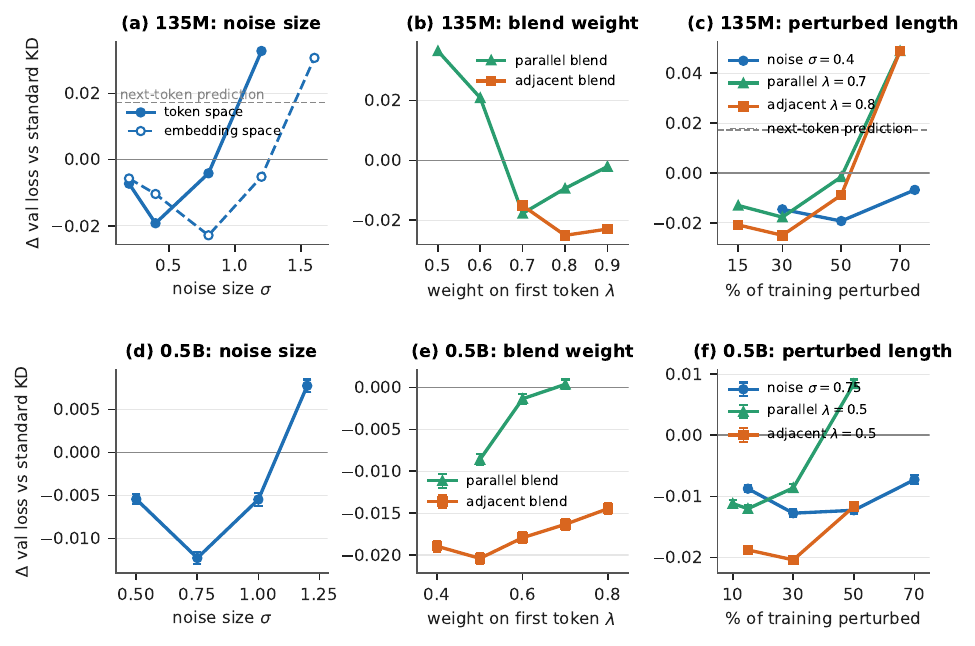}
\caption{\textbf{Change in final validation loss against standard distillation} (below zero is
better). (a--c) 135M student with the SmolLM2-1.7B teacher, 4.2B tokens. The
dashed line is next-token prediction without a teacher. (a) Noise size, with a 50\% perturbed stage.
The open markers in (a) are Gaussian noise added to each
model's input embedding; their $\sigma$ is relative to the embedding's root-mean-square value, so it is
not in the same units as token noise. (b) Blend weight for both blends, with a 30\% perturbed stage. (c) Length of the perturbed stage at each family's best size. (d--f) 0.5B student with the
Qwen2.5-1.5B teacher, 10B tokens, scored on the full validation split. Error bars are $\pm2$ paired
standard errors. (d) Noise size and (e) blend weight, with the same perturbed-stage lengths as in (a, b).
(f) Length of the perturbed stage at each family's best size.}
\label{fig:results}
\end{figure}

\paragraph{All three perturbations improve on standard distillation at 135M
(Fig.~\ref{fig:results}a--b).} For scale, distillation itself improves on next-token prediction by 0.017
nats. Adjacent blending is the strongest variant. At $\lambda{=}0.8$ it lowers the loss by a further
0.025 nats, almost one and a half times what distillation itself gains. Noise at $\sigma{=}0.4$ and
parallel blending at $\lambda{=}0.7$ gain 0.019 and 0.018. Every family has a clear optimum, and going
well past it hurts: large noise, and parallel blends heavier than $\lambda{=}0.7$, end worse than standard
distillation. Gaussian noise added to the embeddings behaves the same way, with its optimum at a somewhat
larger $\sigma$ in its own units.

\paragraph{Length of the perturbed stage (Fig.~\ref{fig:results}c, f).} For both blends at 135M, a
perturbed stage of 30\% of training works best. A shorter stage gives less of the gain, and when 70\% of
training is perturbed both blends end up worse than next-token prediction. Noise is more forgiving. It
does best with half of training perturbed and still helps at three quarters. Blended inputs are much
further from natural text than small noise, which is presumably why less training on them is useful.
The 0.5B runs show the same ordering. Adjacent blending is again best with a 30\% stage, noise does about
equally well with 30\% or 50\%, and parallel blending, the heaviest perturbation we use, does best with
only 15\% and is harmful at 50\%.

\paragraph{The gains hold at 0.5B (Fig.~\ref{fig:results}d--e).} With the Qwen2.5-1.5B teacher,
adjacent blending is again the strongest variant. An even blend of the two tokens works best and improves
on standard distillation by 0.020 nats, with a paired standard error of 0.0003. Two more initialization
seeds of this run and its baseline gave 0.021 and 0.019. The gain shrinks steadily as the blend becomes
less even. Noise is best at $\sigma{=}0.75$, where it gains 0.012 nats, and parallel blending is again
best with an even blend, where it gains 0.009 with a 30\% stage and 0.012 with its best, 15\% stage. Its gain is gone by $\lambda{=}0.7$, the best parallel
blend at 135M. Since the two documents play symmetric roles, $\lambda{=}0.5$ is the heaviest blend this
family allows.

\paragraph{Too little or too much noise.} At both scales the gain from noise first rises and then falls
with its size. Small noise barely changes the teacher's predictions. At 0.5B, $\sigma{=}0.4$ raises the
frozen teacher's loss by only 0.07 nats, and $\sigma{=}0.5$ gives less than half the gain of
$\sigma{=}0.75$. The teacher's targets on such inputs are close to its targets on clean text, so the
student gets little that standard distillation does not already provide. Large noise hurts. At
$\sigma{=}1.2$ the teacher's loss rises by 3.5 nats, and the student ends up worse than with standard
distillation. The useful range lies in between, where the perturbation changes the teacher's predictions
without making them uninformative. We do not have a principled way to locate this range in advance. The
best size differs between our two settings, $\sigma{=}0.4$ at 135M and $0.75$ at 0.5B, and we found it by
sweeping.

\section{Ablations}
\label{sec:ablation}

\begin{figure}[t]
\centering
\includegraphics[width=0.92\linewidth]{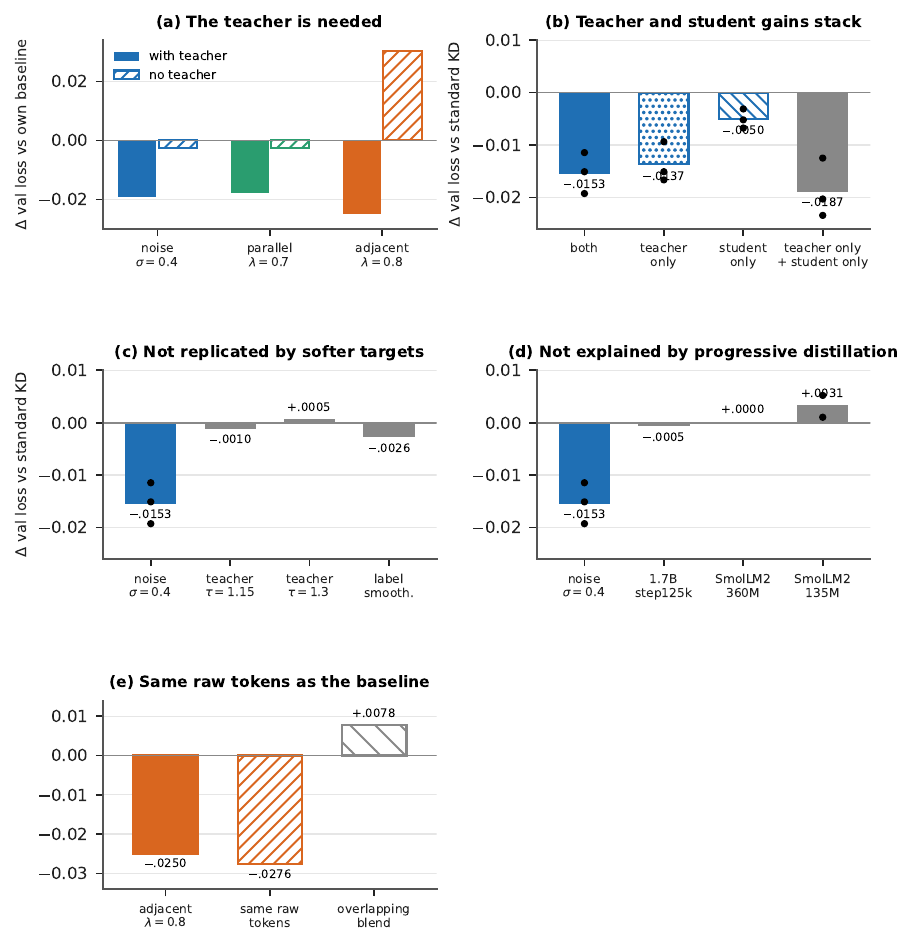}
\caption{\textbf{Ablations} (135M, SmolLM2-1.7B teacher; difference in final validation loss, lower is
better). Bars show the mean over initialization seeds and dots the individual seeds, each compared
with standard distillation from the same seed. (a) Each family at its best setting, with and without the
teacher, each compared with its own unperturbed baseline (standard distillation or next-token
prediction). (b) Token noise given to both models, to the teacher only, or to the student only. The grey
bar is the sum of the two one-sided gains. (c) Clean inputs, with the teacher's target softened instead,
either by a temperature $\tau$ ($\tau{=}1.15$ matches the entropy of the teacher's prediction on
$\sigma{=}0.4$ inputs) or by label smoothing to the same entropy. (d) Clean inputs
for the first half of training, distilled from a weaker model (an early checkpoint of the teacher, or a
smaller model) instead of the teacher. (e) Adjacent blending at $\lambda{=}0.8$, as in the main runs and
on the same raw tokens as the baseline, and the overlapping variant described in the text (one seed).}
\label{fig:ablation}
\end{figure}

The ablations ask where the gain comes from (Fig.~\ref{fig:ablation}). For our perturbations, the
answer is the teacher's predictions on perturbed inputs. The gain needs a teacher, and it comes mostly from perturbing the
teacher's input rather than the student's. It is not explained by the teacher's targets simply being
softer or worse. It also does not depend on two details of our construction, sharing one draw between the
two models and perturbing in token space. Nor, for adjacent blending, does it come from the extra text it reads. These
ablations identify important factors rather than a complete explanation of the gains. All ablations are at 135M and, unless noted, use token noise with $\sigma{=}0.4$ for the first half of training, the best
noise setting. Its gain is 0.019 nats for the seed shown in Fig.~\ref{fig:results} and 0.015 averaged
over three seeds. Where we have several initialization seeds, bars show the mean and dots the individual
seeds, and each seed is compared with standard distillation from the same seed. The gain from noise
varies by up to 0.008 nats between seeds, which matters when reading the smaller differences below.

\paragraph{Without a teacher (Fig.~\ref{fig:ablation}a).} Running the same noise or parallel-blend
schedule with ordinary next-token targets and no teacher improves on next-token prediction by only 0.003
nats, about a seventh of the gain with a teacher. For adjacent blending the contrast is starker. Without a
teacher, training on the blended inputs ends 0.03 nats behind plain next-token prediction, although the
same inputs give the largest gain of all when a teacher labels them.

\paragraph{Teacher side and student side (Fig.~\ref{fig:ablation}b).} Next we perturb only one model's
input. When only the teacher's input is perturbed, the student sees clean text but is trained toward the
teacher's prediction on a perturbed version of it. Averaged over three seeds, this recovers about 90\% of
the gain. When only the student's input is perturbed, with the teacher's prediction on the clean text as
the target, the gain falls to about a third, not much more than the same noise gives without any
teacher. This student-only arm is the usual form of data augmentation for language models, which
perturbs the input and keeps the target \citep{xu2026dataconstrained,chen2026augmentation}. The two
one-sided gains add up to about a fifth more than the gain with both sides perturbed, so the two effects
combine roughly additively. Appendix~\ref{app:rep} compares random-token replacement, the augmentation used in concurrent work on data-constrained pretraining.

\paragraph{Softer targets (Fig.~\ref{fig:ablation}c).} A perturbed input makes the teacher less certain.
At $\sigma{=}0.4$ the entropy of its prediction rises from 1.97 to 2.46 nats, so the gain might come from
a flatter target alone. It does not. Softening the teacher's prediction on clean inputs to the same
entropy, with a temperature or with label smoothing, gains at most 0.003 nats, and a stronger temperature
gains nothing. Applying the usual distillation temperature to both models, which reweights the loss
without changing its optimum, is slightly harmful.

\paragraph{A weaker early teacher (Fig.~\ref{fig:ablation}d).} A perturbed teacher is also a worse
predictor, and distilling first from a weaker or partly trained teacher can act as a curriculum
\citep{panigrahi2025progressive}. To test this, we distill the first half of training on clean inputs from
a weaker model instead: SmolLM2-135M, whose loss is close to that of the 1.7B teacher on $\sigma{=}0.4$
inputs, SmolLM2-360M, or the earliest published checkpoint of the 1.7B teacher. None of them helps. All
three end within 0.003 nats of standard distillation.

\paragraph{Shared draws and token space are not required.} Drawing the teacher's and the student's
noise independently gives the same gain as a shared draw (Appendix~\ref{app:mech}), and Gaussian noise
added to each model's input embedding reaches a similar gain at its own best size
(Fig.~\ref{fig:results}a). The one-hot construction is a convenient way to define one perturbation
for models of different width, but the gain does not come from it. What the method needs is a teacher
that predicts on the perturbed input. Relatedly, in teacher-only experiments, averaging four noisy teacher
predictions gives a smaller gain than using one draw (Appendix~\ref{app:mech}). A plausible explanation is
that noise encourages the student to focus on what the teacher predicts consistently across perturbed
inputs rather than fitting details that change from one perturbation to another, and averaging reduces
that variability without changing the expected target.

\paragraph{The extra text is not the source of adjacent blending's gain (Fig.~\ref{fig:ablation}e).} Because
adjacent blending packs two corpus tokens into each input position, its perturbed stage reads twice as much
text as standard distillation does in the same steps. To remove this difference, we rerun adjacent blending
at $\lambda{=}0.8$ on exactly the tokens the baseline reads in that stage, packed into half as many
positions, so the blended student sees no extra text and also gets less training signal per step. On the
same seed it gains 0.028 nats, slightly more than the 0.025 of the regular run. We also tried an overlapping variant, in which
each position blends a token with the one after it, so that the student again reads exactly the baseline's
text at every position. It ends 0.008 nats worse than standard distillation, but it changes the task as well,
because each input then contains part of the token the model is asked to predict.

\section{Data-constrained pretraining}
\label{sec:dc}

\begin{figure}[t]
\centering
\includegraphics[width=0.9\linewidth]{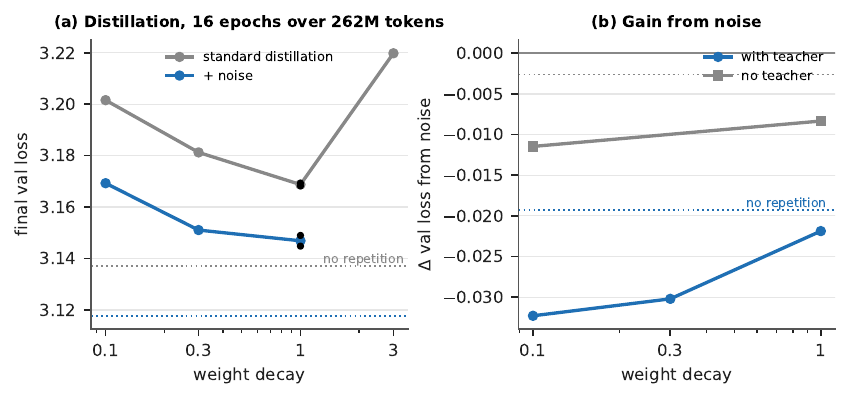}
\caption{\textbf{Data-constrained pretraining} at 135M: 16 epochs over 262M unique tokens, same 4.2B-token
budget. (a) Final validation loss of standard distillation with and without noise (token noise
$\sigma{=}0.4$ for the first 50\% of training) as a function of AdamW weight decay. Black dots are two
initialization seeds at weight decay 1, and dotted lines show the same 4.2B-token budget without repeated
data (weight decay 0.1). Weight decay 10 diverges. (b) The gain from the same noise with and without a
teacher. Dotted lines again show the gain without repeated data. Exact values in Table~\ref{tab:dc} (Appendix~\ref{app:dc}).}
\label{fig:dc}
\end{figure}

When unique data are limited, training repeats them for several epochs, and regularization becomes
important: properly tuned weight decay changes conclusions \citep{kim2026infinite,xu2026dataconstrained},
and input corruption with clean targets helps next-token training
\citep{xu2026dataconstrained,chen2026augmentation,ni2025superdata}. We test whether the gain of
distilling on perturbed inputs survives this setting and a tuned baseline. At 135M, with the same 4.2B-token
budget, training repeats the first 262M tokens of the corpus for 16 epochs. The tokens are cut into
non-overlapping 1024-token windows, reshuffled each epoch with the same shuffle for every run. We sweep AdamW weight decay over 0.1, 0.3, 1, 3 and 10 for standard distillation, of which 10 diverges,
and use token noise $\sigma{=}0.4$ for the first 50\% of training (Fig.~\ref{fig:dc}).

\paragraph{Gain on a tuned baseline.} Weight decay matters under repetition. Raising it from the default
of 0.1 to 1.0, the best value for standard distillation, lowers that baseline's loss by 0.033 nats. On
this tuned baseline, distilling on perturbed inputs still gains 0.023 and 0.020 nats for the two seeds,
while standard distillation itself differs by only 0.001 between them, and noise does not shift the best
weight decay. The same noise without a teacher gains 0.008, so the teacher makes the gain about 2.6 times
larger. At the default weight decay the gain from noise looks larger, 0.032, but much of that is
regularization that a tuned weight decay also provides. On the tuned baseline the gain is about the same
as without any repetition: with fresh data and the same weight decay of 1.0, noise gains 0.020 nats. A
weight decay of 1.0 also helps standard distillation on fresh data, by 0.005 nats, so our default
baseline is slightly under-tuned, but the gains do not shrink on the tuned one. Adjacent blending, in a
single run, gains 0.035 there, compared with 0.025 at the default weight decay. With milder repetition, 4 epochs over 1.05B unique tokens, repetition costs
standard distillation only 0.006 nats and the gain is 0.023.

\section{Quantization and downstream evaluation}
\label{sec:analysis}

We evaluate the final students, standard distillation and each family's best setting, after
post-training weight quantization at 0.5B (Table~\ref{tab:quant}) and on zero-shot benchmarks at both scales (Table~\ref{tab:downstream}). Quantization
damage is the increase in validation loss on 512 fixed validation windows, paired across models; all
linear layers except the output head are quantized to int4 or int3 with groups of 128 weights. We
compare round-to-nearest (RTN, symmetric absmax scale per group) with GPTQ \citep{frantar2023gptq}
(symmetric) and AWQ \citep{lin2024awq} (asymmetric) as implemented in llm-compressor 0.14, both
calibrated on 256 windows of 1024 tokens from each model's own training data.

\begin{table}[t]
\centering
\caption{\textbf{Increase in validation loss from post-training weight quantization} at 0.5B (nats; 512
fixed validation windows, paired across models). RTN: round-to-nearest; GPTQ and AWQ as implemented in
llm-compressor. All linear layers except the output head are quantized; activations stay in 16 bits. Perturbed stage: 30\%, or 15\% for the parallel blend. Bold:
better than standard KD.}
\label{tab:quant}
\vspace{4pt}
\small
\begin{tabular}{lc|ccc|ccc}
\toprule
 & full & \multicolumn{3}{c|}{int4, group 128} & \multicolumn{3}{c}{int3, group 128} \\
 & precision & RTN & GPTQ & AWQ & RTN & GPTQ & AWQ \\
\midrule
standard KD & 2.6782 & 0.205 & 0.028 & 0.019 & 6.226 & 0.160 & 0.099 \\
noise $\sigma{=}0.75$ & \textbf{2.6642} & 0.317 & \textbf{0.027} & 0.019 & \textbf{5.623} & \textbf{0.117} & \textbf{0.097} \\
parallel $\lambda{=}0.5$ & \textbf{2.6660} & \textbf{0.161} & \textbf{0.024} & 0.020 & \textbf{1.480} & \textbf{0.123} & 0.099 \\
adjacent $\lambda{=}0.5$ & \textbf{2.6579} & \textbf{0.113} & \textbf{0.024} & 0.019 & \textbf{1.203} & \textbf{0.120} & \textbf{0.098} \\
\bottomrule
\end{tabular}

\end{table}

\begin{table}[t]
\centering
\caption{\textbf{Zero-shot accuracy} (\%, lm-evaluation-harness; normalized accuracy for HellaSwag, ARC,
PIQA and SciQ, accuracy for Winogrande and LAMBADA). Each family at its best setting for that scale, with a perturbed stage of 30\% (0.5B parallel blend
15\%, 135M noise 50\%). $\Delta$
Avg.: difference from standard distillation in the average over tasks, $\pm$ paired bootstrap
standard error over examples. Best value per scale in bold.}
\label{tab:downstream}
\vspace{4pt}
\small
\setlength{\tabcolsep}{4pt}
\begin{tabular}{lccccccc|cc}
\toprule
 & HellaSwag & ARC-E & ARC-C & PIQA & Wino. & LAMBADA & SciQ & Avg. & $\Delta$ Avg. \\
\midrule
\multicolumn{10}{l}{\textit{0.5B}} \\
standard KD & 45.0 & 57.9 & 29.0 & 68.0 & 52.7 & 37.5 & 77.3 & 52.5 & -- \\
noise $\sigma{=}0.75$ & 45.7 & 56.8 & 30.5 & 68.8 & 52.8 & \textbf{39.1} & \textbf{81.3} & \textbf{53.6} & $+1.1\pm0.4$ \\
parallel $\lambda{=}0.5$ & 45.5 & \textbf{58.1} & 30.3 & 68.3 & \textbf{54.1} & 36.7 & 77.5 & 52.9 & $+0.5\pm0.3$ \\
adjacent $\lambda{=}0.5$ & \textbf{45.9} & 58.0 & \textbf{30.6} & \textbf{69.7} & 53.4 & 36.3 & 80.9 & 53.5 & $+1.1\pm0.3$ \\
\midrule
\multicolumn{10}{l}{\textit{135M}} \\
standard KD & 30.1 & 44.4 & 25.2 & 60.1 & \textbf{52.3} & 21.3 & 64.0 & 42.5 & -- \\
noise $\sigma{=}0.4$ & 30.4 & 45.4 & 24.5 & \textbf{61.3} & 50.2 & \textbf{22.6} & \textbf{65.1} & 42.8 & $+0.3\pm0.3$ \\
parallel $\lambda{=}0.7$ & \textbf{30.5} & 46.1 & 25.1 & 60.3 & 50.0 & 22.1 & 62.6 & 42.4 & $-0.1\pm0.4$ \\
adjacent $\lambda{=}0.8$ & 30.4 & \textbf{46.8} & \textbf{25.3} & 60.9 & 50.8 & 22.3 & \textbf{65.1} & \textbf{43.1} & $+0.6\pm0.3$ \\
\bottomrule
\end{tabular}

\end{table}

\paragraph{The gains survive quantization.} With GPTQ at int4, every perturbed student keeps its
advantage over standard distillation quantized the same way. Adjacent blending's advantage even grows
slightly, from 0.020 to 0.024 nats, and those of parallel blending and noise also grow slightly, to 0.017 and 0.015.
With AWQ at int4 the advantages are essentially unchanged. So with either method the perturbed students
quantize at least as well as the standard one. Plain round-to-nearest is mixed: at int4 the noise-trained
student loses more than standard distillation, and the blended students lose less.

\paragraph{Round-to-nearest and weight kurtosis.} At int3 under round-to-nearest the blended students
are far more robust. Int3 raises the loss of standard distillation by 6.2 nats, but that of the two blends
by only 1.2 and 1.5. The difference persists when we compare checkpoints from the middle of training at
matched full-precision loss, and it tracks the weight distribution. The mean excess kurtosis of the
linear-layer weights is about 0.5 for standard distillation and for noise, but 0.14 for both blends.
Parallel blending at $\lambda{=}0.7$, which gives no gain at 0.5B, keeps the standard student's kurtosis
and its round-to-nearest damage. Methods that compensate for quantization error remove most of the
difference. With GPTQ the blended students' int3 damage is a quarter lower than standard distillation's,
similar to noise, and with AWQ all students lose the same. The activation statistics we checked, the
outlier ratio of the inputs to each linear layer and the fraction of very large residual-stream
activations, do not differ between the students.

\paragraph{Downstream accuracy.} Averaged over seven zero-shot tasks from lm-evaluation-harness,
adjacent blending and noise both raise accuracy at 0.5B by about one point over standard distillation,
from 52.5\% to 53.5\% and 53.6\% (Table~\ref{tab:downstream}). Parallel blending gains half a point. The
largest gains are on SciQ, about four points, and HellaSwag, about one, while both blends lose about a
point on LAMBADA. At 135M the differences in the average are small, at most 0.6 points.

\section{Related work}

\paragraph{Distillation in pretraining.} Distillation into small pretrained language models is
well established \citep{hinton2015distilling,peng2025pretraining}. Progressive distillation from
intermediate teacher checkpoints can induce an implicit curriculum \citep{panigrahi2025progressive}; a
weaker first-stage teacher does not reproduce our gain (Section~\ref{sec:ablation}). Work on language-model
distillation has mostly changed the divergence or the sampling distribution, for example with
on-policy or reverse-KL objectives \citep{gu2024minillm,agarwal2024gkd}. MiniPLM \citep{gu2025miniplm} uses the teacher to reweight the pretraining data, and
\citet{busbridge2025distillation} give scaling laws for distillation. We keep the objective
fixed and change the inputs on which it is evaluated.

\paragraph{Perturbations and derivative matching.} Matching a teacher's derivatives, or its
responses to perturbations, goes back to Jacobian and Sobolev training
\citep{srinivas2018jacobian,czarnecki2017sobolev}. In vision, \citet{beyer2022patient} show that
teacher and student should see the same augmented view. Task-specific distillation of language models already trains on teacher-labeled augmented
text, e.g.\ with word replacement \citep{jiao2020tinybert} or adversarially masked text
\citep{rashid2021matekd}, and \citet{liu2021noisy} inject noise into both teacher and student in
self-distillation for summarization. In vision, teacher-labeled augmentation is analyzed by
\citet{wang2022goodaug} and \citet{stanton2021kd}; Noisy Student \citep{xie2020noisystudent} instead
noises the student and labels the clean input. We study the pretraining setting and separate the
teacher-side and student-side contributions. Unlike Jacobian matching,
our students do not learn the teacher's local response to the perturbation.

\paragraph{Data-constrained pretraining and input augmentation.} Repeating data for a few epochs is
nearly as good as fresh data, with returns diminishing beyond about 16 epochs
\citep{muennighoff2023scaling,chai2026repeated}. Under repetition, dropout and much larger weight decay
help \citep{xue2023repeat,kim2026infinite}, and so does corrupting the input while keeping the clean
target: an auxiliary loss on masked inputs \citep{xu2026dataconstrained}, random token replacement
\citep{chen2026augmentation}, or masking, which partly explains why diffusion language models tolerate
more repetition \citep{ni2025superdata,prabhudesai2025diffusion}. \citet{kim2026infinite} also distill an
ensemble into a smaller student, by training on the ensemble's generated text. These methods perturb the
student's input only; in distillation, we find that most of the benefit comes from letting the teacher
predict on the perturbed input.

\paragraph{Mixing and merging inputs.} Mixup-style interpolation \citep{zhang2018mixup} has been
applied to language-model distillation in embedding space \citep{liang2021mixkd}. Merging
neighboring tokens is used to save computation in vision transformers \citep{bolya2023tome} and
in language-model pretraining \citep{peng2026tst}, and embedding noise improves instruction
fine-tuning \citep{jain2023neftune}. Our blends use related constructions as shared
perturbations for distillation, and our ablation shows that their benefit depends on the teacher.

\section{Conclusion}

Distilling on perturbed inputs is a small change to standard distillation. Only the inputs of an early
stage of training change, and the objective, the data and the token budget stay the same. It improved the
student at both scales we tested, by 0.025 nats at 135M and 0.020 nats at 0.5B with the best variant,
adjacent-token blending. In our experiments the perturbed stage worked best when it was short, 15--50\% of training
depending on how strongly the inputs are perturbed. For additive noise,
the improvement comes mostly from the teacher's predictions on the perturbed inputs. This is why it mostly
disappears without a teacher, and why perturbing only the student's input recovers only part of it.

An important limitation is the scale of our experiments: whether these gains persist for larger models or
with substantially more training data remains open. The perturbations studied here provide a proof of
concept that a teacher's predictions beyond ordinary training text can offer useful supervision. A
promising direction for future work is to query the teacher on other inputs beyond standard text and the
perturbations discussed here, and to explore how to choose those inputs to improve student learning.

\bibliography{iclr2027_conference}
\bibliographystyle{iclr2027_conference}

\appendix
\section{Training details}
\label{app:details}

\begin{table}[h]
\centering
\caption{\textbf{Training configuration.} Both scales train the student from random initialization with
a frozen teacher; the distillation loss is the forward KL divergence over the full vocabulary at
temperature 1, with no next-token term.}
\vspace{4pt}
\small
\setlength{\tabcolsep}{4.5pt}
\begin{tabular}{lll}
\toprule
 & 135M & 0.5B \\
\midrule
Student architecture & SmolLM2-135M & Qwen2.5-0.5B \\
Teacher (frozen) & SmolLM2-1.7B & Qwen2.5-1.5B \\
Vocabulary (shared) & 49{,}152 & 151{,}936 \\
Data & FineWeb-Edu, 58.6B-token pool & FineWeb-Edu, 65B-token pool \\
Steps $\times$ batch $\times$ length & $8000\times512\times1024$ (4.2B tokens) & $19073\times256\times2048$ (10B tokens) \\
Optimizer & AdamW, $\beta{=}(0.9,0.95)$, weight decay 0.1 & same \\
Peak learning rate & $6\times10^{-4}$ & $1.2\times10^{-3}$ \\
Schedule & WSD: 10\% warmup, 10\% decay & same \\
Precision & bfloat16 & bfloat16 \\
Compute per run & ${\approx}20$ H100-hours & ${\approx}90$ H100-hours \\
\bottomrule
\end{tabular}
\end{table}

\section{Data-constrained results}
\label{app:dc}

\begin{table}[h]
\centering
\caption{\textbf{Data-constrained pretraining} at 135M: final validation loss with 16 epochs over 262M
unique tokens, by weight decay, against the same 4.2B-token budget without repetition (1 epoch). Two
values: two initialization seeds. The 1-epoch runs at weight decay 1 use one seed. Noise: token noise $\sigma{=}0.4$ for the first 50\% of training. Bold:
the better of each objective with and without noise.}
\label{tab:dc}
\vspace{4pt}
\small
\begin{tabular}{lcccccc}
\toprule
 & \multicolumn{2}{c}{1 epoch, wd} & \multicolumn{4}{c}{16 epochs, weight decay} \\
\cmidrule(lr){2-3}\cmidrule(lr){4-7}
 & 0.1 & 1 & 0.1 & 0.3 & 1 & 3 \\
\midrule
Next-token prediction & 3.1542 & -- & 3.3365 & -- & 3.2297 & -- \\
\quad + noise & \textbf{3.1516} & -- & \textbf{3.3250} & -- & \textbf{3.2214} & -- \\
\midrule
Standard distillation & 3.1370 & 3.1318 & 3.2016 & 3.1812 & 3.1682 / 3.1693 & 3.2198 \\
\quad + noise & \textbf{3.1177} & \textbf{3.1120} & \textbf{3.1693} & \textbf{3.1510} & \textbf{3.1448 / 3.1490} & -- \\
\bottomrule
\end{tabular}

\end{table}

\section{Augmentation from concurrent work}
\label{app:rep}

\begin{figure}[h]
\centering
\includegraphics[width=0.5\linewidth]{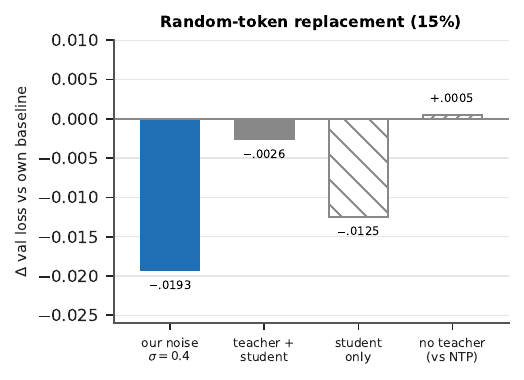}
\caption{\textbf{Random-token replacement} (135M, one seed). Replacing 15\% of tokens with random ones
\citep{chen2026augmentation}, given to both models, to the student only, or without a teacher, next to
token noise on the same seed. Each arm is compared with its own baseline (standard distillation, or
next-token prediction for the arm without a teacher).}
\label{fig:rep}
\end{figure}

Recent work on
data-constrained pretraining perturbs the student's input and keeps the clean target. Its most effective
variant replaces a random 15\% of tokens with random vocabulary items \citep{chen2026augmentation}. We
tried this replacement inside distillation in the same three ways, with one seed. Without a teacher it
does not improve next-token training in our setting. With a teacher it behaves differently from our
perturbations. Applied to the student's input only, with the teacher predicting on clean text, it gains
0.013 nats, while giving the replaced text to both models gains only 0.003. A replaced token is gone from
the input, so the teacher's prediction on it is a prediction about different text, whereas our
perturbations keep the true token. Student-side replacement still falls short of noise and adjacent
blending on the same seed, which gain 0.019 and 0.025. We did not test the masked auxiliary objective of
\citet{xu2026dataconstrained}, which needs a second forward pass at every step. Fig.~\ref{fig:rep} shows the results.

\section{Mechanism probes}
\label{app:mech}

\begin{figure}[h]
\centering
\includegraphics[width=0.6\linewidth]{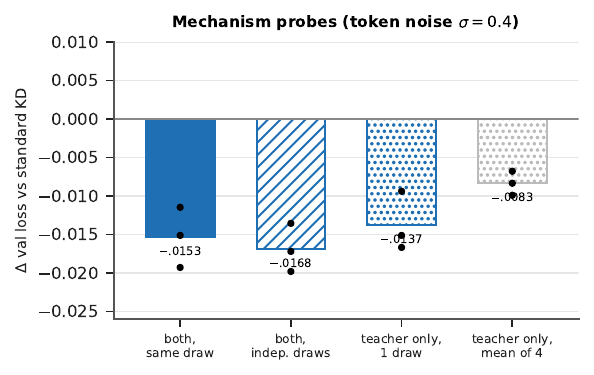}
\caption{\textbf{Mechanism probes} (135M, token noise $\sigma{=}0.4$ for the first 50\% of training;
difference from standard distillation, mean over seeds with individual seeds as dots).}
\label{fig:mech}
\end{figure}

\emph{Shared or independent draws.} Drawing the teacher's and the student's noise independently gives the
same gain as sharing it ($-0.0168$ against $-0.0153$, three seeds; Fig.~\ref{fig:mech}). In line with this, the students do not learn how the
teacher's prediction responds to the perturbation. We measure each model's response as the change in its
next-token distribution under a small step along the noise direction. For every student we trained, this
response is nearly uncorrelated with the teacher's (correlation at most about 0.03), with or without noise in
training. \emph{One draw or the average of four.} Averaging the teacher's predictions
over four noise draws into one target, which has the same expected value as a single draw, gives a
smaller gain than a single draw given to the teacher only ($-0.0076$ against $-0.0122$ on the same two seeds). This suggests
that the randomness of the target helps, not only its expected value. The differences in this appendix are comparable to the spread between
seeds, and we regard them as indicative.

\end{document}